\documentclass[10pt,journal]{IEEEtranTCOM}
\usepackage{graphicx} 
\usepackage[english]{babel}
\usepackage{amsthm}
\usepackage{amsmath}
\usepackage{amsfonts}
\usepackage{mathtools}
\usepackage{booktabs}
\usepackage{url}
\usepackage{geometry}
\usepackage[section]{placeins}
\usepackage{amssymb}

\newcommand{\be}{\begin{equation}}
\newcommand{\ee}{\end{equation}}

\usepackage[hidelinks]{hyperref}
\usepackage{algorithm}
\usepackage{algorithmic}

\usepackage{amsmath}
\usepackage{tikz}

\title{Linear cost mutual information estimation and independence test of similar performance as HSIC}
\author{
\IEEEauthorblockN{Jarek Duda, Jagiellonian University, Cracow, Poland, \emph{dudajar@gmail.com}}\\
\IEEEauthorblockN{Jagoda Bracha, University of Warsaw, Warsaw, Poland}\\
\IEEEauthorblockN{Adrian Przybysz, Collegium Da Vinci in Poznan, Poland}}
\begin{document}
\maketitle
\begin{abstract} 
Evaluation of statistical dependencies between two data samples is a basic problem of data science/machine learning, and HSIC (Hilbert-Schmidt Information Criterion)~\cite{HSIC} is considered the state-of-art method. However, for size $n$ data sample it requires multiplication of $n\times n$ matrices, what currently needs $\sim O(n^{2.37})$ computational complexity~\cite{mult}, making it impractical for large data samples. We discuss HCR (Hierarchical Correlation Reconstruction) as its linear cost practical alternative, in tests of even higher sensitivity to dependencies, and additionally providing actual joint distribution model for chosen significance level, by description of dependencies through features being mixed moments, starting with correlation and homoscedasticity. Also allowing to approximate mutual information as just sum of squares of such nontrivial mixed moments between two data samples. Such single dependence describing feature is calculated in $O(n)$ linear time. Their number to test varies with dimension $d$ - requiring $O(d^2)$ for pairwise dependencies, $O(d^3)$ if wanting to also consider more subtle triplewise, and so on.
\end{abstract}
\section{Introduction}
\begin{figure}[t!]
    \centering
        \includegraphics[width=90mm]{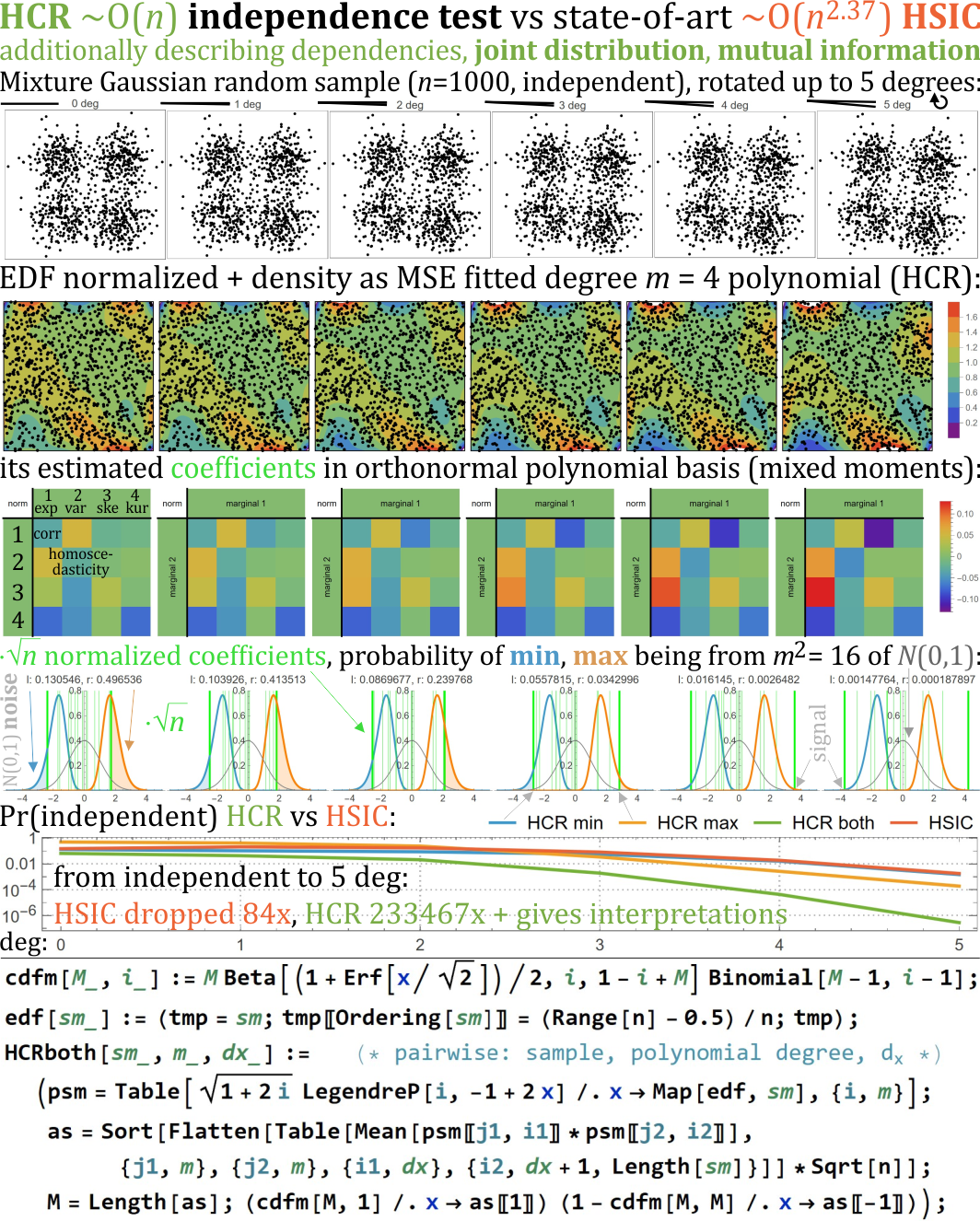}
        \caption{Proposed HCR independence test and comparison with state-of-art HSIC of much higher complexity, impractical for large samples, which are required to find very subtle dependencies. We generate shown $n=1000$ size random dataset from Gaussian mixtures independently for both coordinates, and slightly rotate it to introduce dependence. In HCR we first separately normalize coordinates to nearly uniform distribution in $[0,1]$ by CDF/EDF, then model joint density as a chosen degree polynomial, here $m=4$. Using orthonormal basis of Legendre polynomials, coefficients are approximately: expected value, variance, skewness, kurtosis, and their dependencies. Sum of squares of shown $m^2$ nontrivial coefficients approximate mutual information, and allows independence test: by first normalizing multiplying by $\sqrt{n}$, and then testing if they agree with $N(0,1)$ set of variables, where the presented test of extreme coefficients has turned out the most sensitive to dependencies, also evaluating  significance of contributions to joint distribution. There is shown probability of exceeding these values with min/max from $m^2=16$ independent $N(0,1)$ variables, also both by taking their product using shown Mathematica code, properly indicating dependence with much higher certainty than HSIC, additionally providing its description with mixed moments and model of joint density.  }
        \label{intr}
\end{figure}

Statistical dependencies start with correlation coefficients, defining dependence between expected values of two random variable: their first moments. There are also higher moments, like variance, skewness, kurtosis - which also contribute to statistical dependencies, like homoscedasticity evaluating co-occurrence of high variance common e.g. in financial time series or Fig. \ref{intr}.

Hierarchical Correlation Reconstruction (HCR, \cite{HCR,HCRNN})  \textbf{automatically decomposes dependencies into such mixed moments} of two or more variables. Specifically, like in copula theory~\cite{copula}, we start with normalization of all variables to nearly uniform distribution in $[0,1]$ ($\sim$ quantiles) by $\hat{x}= \textrm{CDF}(x)$ or EDF (cumulative/empirical distribution function), separately for each coordinate. Then we model joint density of such normalized variables as a linear combination in product basis: 
\be\rho(\textbf{x})\approx \sum_{\textbf{j}\in B} a_\textbf{j}\, f_{\textbf{j}} (\textbf{x}) = \sum_{(j_1,\ldots,j_d)\in B} a_\textbf{j}\, f_{j_1}(x_1)\cdot\ldots\cdot f_{j_d}(x_d)\ee for bold $\textbf{x}=(x_1,\ldots,x_d)$ denoting dimension $d$ vectors, and $B$ chosen basis of mixed moments. For any continuous joint distribution we can approximate it with polynomials as close as we want, the $a_\textbf{j}$ coefficients become moments. Using orthonormal basis of (Legendre) polynomials $\int_0^1 f_k(x)f_l(x)dx =\delta_{kl}$, coefficients can be MSE \textbf{estimated} as $a_\textbf{j}=\frac{1}{n} \sum_{\textbf{x}} f_\textbf{j}(\textbf{x})$ mean over normalized sample, in $O(n)$ time for $|X|=n$ data sample. 

Thanks to orthonormality, we can approximate mutual information as sum of squares of such nontrivial moments between two variables~\cite{HCRNN}: $I(X,Y)\approx \sum_{\textbf{j},\textbf{k}\neq \textbf{0}} (a_{\textbf{j},\textbf{k}})^2$. As independent variables have \textbf{zero mutual information}, in this article we will adapt it for \textbf{independence test}, asking for probability if all tested nontrivial moments between two variables could be zero. 

Alternative way to test independence is verifying \textbf{if we can factorize joint distribution}: $P_{X,Y}=P_X P_Y$? It is used e.g. in \textbf{HSIC independence test}~\cite{HSIC}, considered as the state-of-art. However, it requires matrix multiplication for $n\times n$ matrices like $K_{ij}=\exp(- \|\textbf{x}^i - \textbf{x}^j\|_2/2\sigma^2)$, and the lowest complexity known algorithm for matrix multiplication has $\sim O(n^{2.37})$ complexity~\cite{mult}, making this test impractical for large data samples.

Therefore, we discuss replacement of $\sim O(n^{2.37})$ HSIC independence test, with practical also for large samples: $O(n |B|)$ cost test based on HCR, for $B$ being a chosen set of features as mixed moments to test, e.g. growing $|B|\sim O(d^2)$ with square of dimension if including only pairwise dependencies, or $O(d^3)$ if adding very subtle triplewise and so on.

Beside much lower cost, making it practical to search very large data samples required to find subtle dependencies, HCR also turns out more sensitive to dependencies. It additionally provides approximation of mutual information, and model of joint distribution with control of significance of contributions.

\begin{figure}[t!]
    \centering
        \includegraphics[width=90mm]{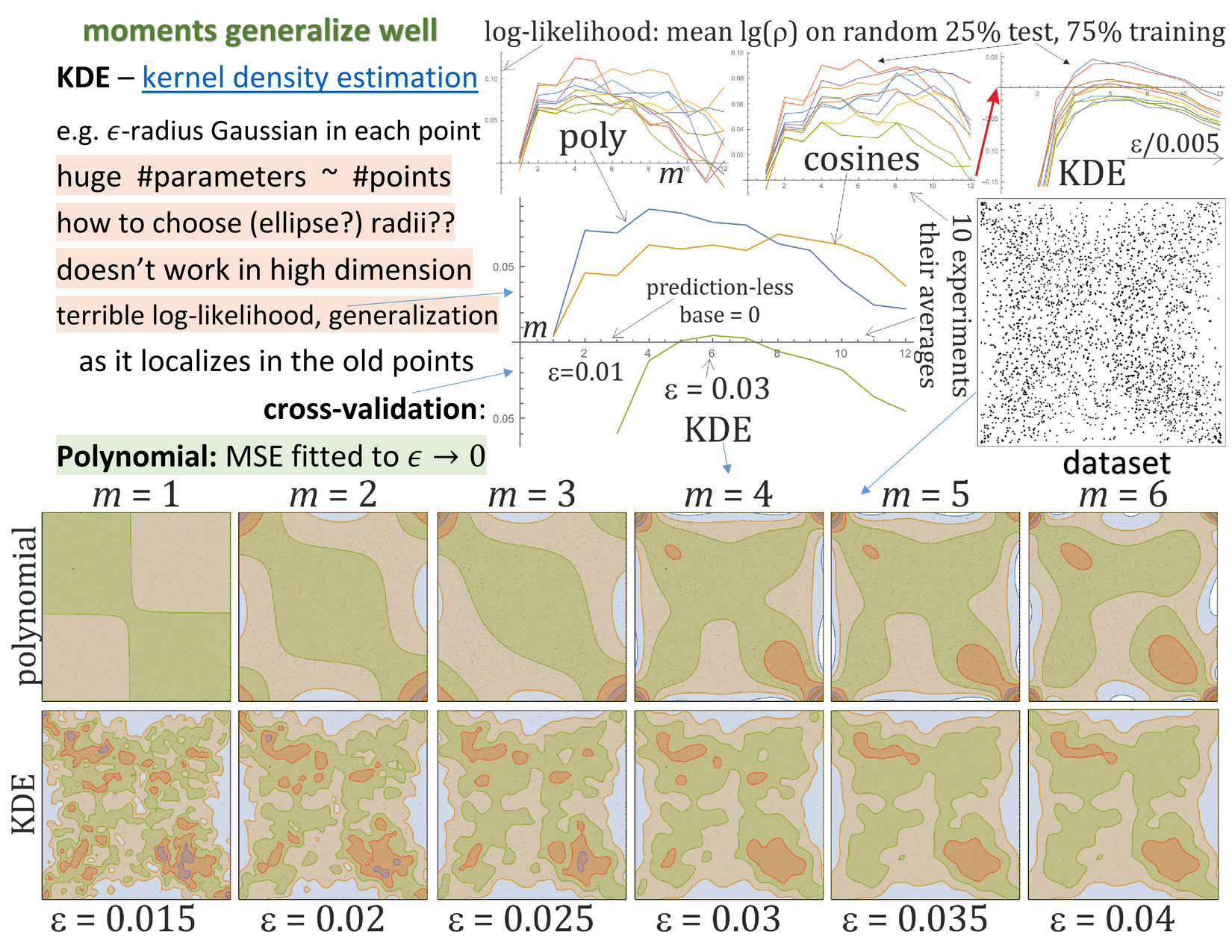}
        \caption{Comparison of joint density estimation using \textbf{global basis} of polynomials/moments we use in HCR, and \textbf{local basis} of Gaussians usually used in HSIC, requiring to choose radius $\epsilon$. Global basis allows to represent the density with a reasonable number of features (moments) e.g. $m^2$ here, allowing  the discussed reduction of complexity e.g. from $\sim O(n^{2.37})$ to $O(m^2 n)$ for $n$ point independence test. Evaluating log-likelihood in cross-validation, local basis barely got above 0 obtained for trivial $\rho=1$ assumption, while global basis finds features as moments, which generalize well from training to test set. Here with the highest log-likelihood for $m=4$ degree up to kurtosis, quite universal e.g. for financial data, we focus on in this article. In contrast, local basis just assumes new points will be close to the old ones, what does not generalize well.     }
        \label{global}
\end{figure}

The original motivation of this article (\cite{HCRNN}) was application for information bottleneck training of neural networks~\cite{IB0}, planned as future work. While originally it required mutual information evaluating the number of bits shared between contents of neural network layers, in practical realizations it was replaced with HSIC~(\cite{IB1,IB2}), emphasizing it is something different, just practical to evaluate. In contrast, for HCR we indeed derive practical formula for approximation of mutual information, as just sum of squares of nontrivial mixed moments between two data samples, also using global basis which is often better for generalization as we can see in Fig. \ref{global}, hopefully leading to improvements of information bottleneck training.

This is early version of article, with main purpose to introduce the method. We plan to extend benchmarks comparing with HSIC library\footnote{HSIC library used for benchmarks: \url{http://pypi.org/project/PyRKHSstats/}} and details in future versions.

\begin{figure}[t!]
    \centering
        \includegraphics[width=90mm]{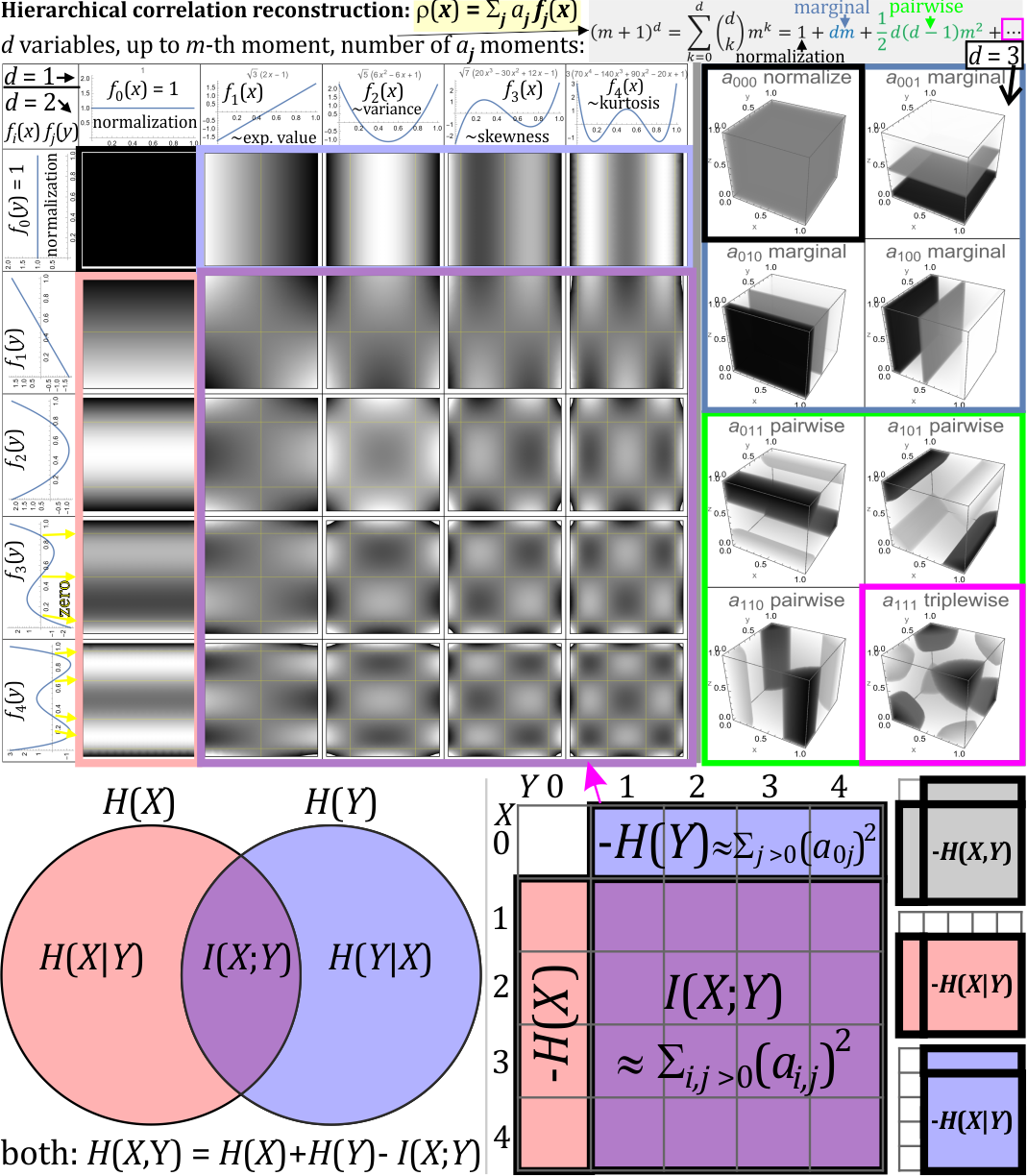}
        \caption{\textbf{Top} left: the first $j=0,1,2,3,4$ from orthonormal polynomial basis $f_j(x)$ for $d=1$ dimension and product basis $f_j(x) f_k(y)$ for $d=2$ dimensions, which linear combination is used as model of (joint) density in HCR, e.g. $\rho(x,y)=\sum_{jk=1}^m a_{jk}f_j(x) f_k(y)$ for variables normalized to nearly uniform distribution in $[0,1]$. As $f_0=1$, top row $a_{i0}$ describes marginal distribution of $X$ and $a_{0j}$ left column of $Y$. Then $i,j\geq 1$ describe their dependencies as mixed moments close to expected value, variance, skewness and kurtosis. Right: $d=3$ case adding much more subtle triplewise dependencies. \\
        \textbf{Bottom} left: standard view on (joint/conditional) entropy of $X,Y$ variables and their mutual information. Right: for HCR using $\ln(1+t)\approx t$ first Taylor term they can be approximated with sums of squares of coefficients, especially mutual information as $I(X;Y)\approx \sum_{i,j=1}^m \left(a_{ij}\right)^2$, also in higher dimensions, which should be close to zero for independent variables.
        }
        \label{basis}
\end{figure}

\section{HCR independence test and evaluation }
Assume we have $X=\{\textbf{x}^i\}_{i=1}^n, Y=\{\textbf{y}^i\}_{i=1}^n$ size $n$ data sample as pairs forming $Z=\{(\textbf{x}^i,\textbf{y}^i)\}_{i=1}^n$ of vectors of $d_x$, $d_y$ dimensions, for which we would like to test independence. To simplify presentation we focus on $d_x=d_y=d$ equal dimension case, but also discuss general situation.

In HCR it is convenient to start with normalization of marginal distributions to nearly uniform in $[0,1]$, then model joint distribution as a linear combination in some orthonormal basis, usually polynomials - which for independent samples should use zero nontrivial moments between them, we can build test from.
\subsection{Normalization to nearly uniform in $[0,1]$ with CDF/EDF}
As in copula theory ~\cite{copula}, we start with normalization of variables to nearly uniform distribution in $[0,1]$ by transforming with CDF/EDF (cumulative or empirical distribution function) to nearly quantiles, independently for each coordinate. 

For CDF approach we need to choose some parametric family e.g. Gaussian, and estimate its parameters e.g. based on dataset, separately for each coordinate. Generally, especially if data does not suit any parametric family, we can use EDF normalization: order values in the size $n$ sample and assign $(i-1/2)/n$ to $i$-th in the order. However, it requires sorting having $O(n \ln n)$ complexity - if linear is required, there should be used CDF.

For the discussed independence test/evaluation in benchmarks we use EDF: for $d$-dimensional data samples $X$ and $Y$, we start with transforming them by applying EDF normalization separately for each coordinate as $\textrm{EDF}^x_i, \textrm{EDF}^y_i$:
\be \hat{Z}=\{(\hat{\textbf{x}},\hat{\textbf{y}}): \hat{x}_i= \textrm{EDF}^x_i(x_i), \hat{y}_i= \textrm{EDF}^y_i(y_i), (\textbf{x},\textbf{y})\in Z\} \ee 

As mutual information is invariant under reparametrization of the marginal variables~\cite{MI}, we can estimate it for such transformed variables instead: $I(X;Y)\approx I(\hat{X};\hat{Y})$.

\subsection{Polynomial density model}
Let us now model joint distribution on $(\textbf{x},\textbf{y})\in \hat{Z}$ normalized variables as a linear combination of usually polynomials $f$:
\be \rho(\textbf{x},\textbf{y})\approx \sum_{\textbf{j}\in B_x, \textbf{k}\in B_y} a_{\textbf{j},\textbf{k}}\, f_\textbf{j} (\textbf{x}) f_\textbf{j}(\textbf{y}) \label{jd}
\ee
for chosen bases: $B_x$,$B_y$ (can be equal), e.g.  $\{0,\ldots,m\}^d$. Using orthonormal family, MSE coefficient estimation becomes:
\be a_{\textbf{j},\textbf{k}} = \textrm{mean}(f_\textbf{j} (\textbf{x}) f_\textbf{k}(\textbf{y})) = \frac{1}{n} \sum_{(\textbf{x},\textbf{y})\in\hat{Z}}  f_\textbf{j} (\textbf{x}) f_\textbf{k}(\textbf{y}) \label{estim}\ee
where we can use product basis $f_\textbf{j} (\textbf{x}) = f_{j_1}(x_1)\cdot \ldots\cdot f_{j_d}(x_d)$ for orthonormal: $\int_0^1 f_k(x)f_l(x)dx =\delta_{kl}$ e.g. polynomials (rescaled Legendre) we will use, getting interpretation close to standard cumulants:
\be f_0 = 1\qquad\textrm{corresponds to normalization}\label{leg}\ee
$$ f_1(x)=\sqrt{3} (2 x-1)\qquad \sim\textrm{expected value}$$
$$f_2(x) =\sqrt{5} \left(6 x^2-6 x+1\right) \qquad \sim\textrm{variance} $$
$$ f_3(x) = \sqrt{7} \left(20 x^3-30 x^2+12 x-1\right)\qquad\sim\textrm{skewness}$$
$$ f_4(x) =3 \left(70 x^4-140 x^3+90 x^2-20 x+1\right)\qquad\sim\textrm{kurtosis} $$

The coefficients can be interpreted as mixed moments, only between nonzero indexes as $f_0=1$, e.g. $a_{00i0}$ is $i$-th moment of 3rd out of 4 variables, $a_{012}$ describes dependence between expected value of 2nd variable, and variance of 3rd variable.

Alternatively there could be used Fourier or DCT basis, especially for periodic variables. As mentioned further, the basis can be automatically optimized e.g. with SVD~\cite{duda2022fast}, for given data sample, or family of data.
\subsection{Entropy, mutual information estimation}
Using $\ln(1+t)\approx t$ first order approximation and orthonormality, we can approximate \textbf{entropy} in nits ($1/\ln(2)\approx 1.44$ bits) from the coefficients:
\be H(\hat{X})=-\int_{[0,1]^d}\rho(\mathbf{x})\,\ln(\rho(\mathbf{x}))\,d\mathbf{u} \approx -\sum_{\mathbf{j}\in B_x^+} (a_\mathbf{j})^2   \ee 
for $B^+ = B\backslash \{\textbf{0}\}$ basis without normalization $f_\textbf{0}=1$. Like visualized in Fig. \ref{basis}, analogously e.g. for joint distribution:
\be H(\hat{X},\hat{Y})\approx -\sum_{(\mathbf{j},\mathbf{k})\in (B_x,B_y)\backslash \{\textbf{0},\textbf{0}\}} (a_{\mathbf{j},\mathbf{k}})^2   \ee 
allowing to approximate \textbf{mutual information} as just sum of squares of nontrivial mixed moments between variables:
\be I(\hat{X},\hat{Y})=H(\hat{X})+H(\hat{Y})-H(\hat{X},\hat{Y}) \approx \sum_{\mathbf{j}\in B_x^+}\  \sum_{\mathbf{k}\in B_y^+} \left(a_{(\mathbf{j},\mathbf{k})}\right)^2 \label{mif} \ee 

However, using these sum of squares formulas with (\ref{estim}) estimator as mean: $a_{\textbf{j},\textbf{k}}=\textrm{mean}(f_\textbf{j} (\textbf{x}) f_\textbf{k}(\textbf{y}))$, it would be artificially increased - let us now discuss it and correction.\\

\paragraph*{Subtracting the variance}
Assume we get $n$ values $\{w^i\}_{i=1..n}$ from $N(\mu,\sigma)$ distribution, and we want to estimate $\mu^2$. A natural approach is calculating the mean $\bar{w}=\frac{1}{n}\sum_i w_i$ and taking $\bar{w}^2$, however, it would be biased (artificially increased).

The Central Limit Theorem (CLT) says that for large sample:
\be\textrm{CLT}:\qquad\qquad \frac{\bar{w}-\mu}{\sigma/\sqrt{n}} \sim N(0,1)\ee
Applying square to both sides, and taking expected values: mean square of value from $N(0,1)$ is 1, hence we get: 
$$ \frac{n}{\sigma^2}\, E[(\bar{w}-\mu)^2]=1$$ 
As $E[\bar{w}]=\mu$, expanding the square we finally get 
\be \mu^2 = E[\bar{w}^2]- \sigma^2/n\ee
suggesting to correct entropy and mutual information evaluation by subtracting the variance in coefficient estimation $(\ref{estim})$:
\be I_c(\hat{X},\hat{Y})=\sum_{\mathbf{j}\in B_x^+}\  \sum_{\mathbf{k}\in B_y^+} \left(\textrm{mean}(f_\textbf{j} (\textbf{x}) f_\textbf{k}(\textbf{y}))^2-\frac{\textrm{var}(f_\textbf{j} (\textbf{x}) f_\textbf{k}(\textbf{y}))}{n}\right) \label{mifc} \ee
we plan to compare in future with state-of-art methods like \cite{MI}.

Also, while we have used only the first term of Taylor series $\ln(1+t)=-\sum_{i=1}^\infty (-t)^i/i$ , it is worth to somehow include higher terms, requiring integrals of 3 and more basis functions.


\subsection{HCR-based independence tests}
While the above formula allows to approximately estimate mutual information $I(X,Y)=I(\hat{X},\hat{Y})$ for $B^+_x\times B^+_y$ chosen basis, for independence test we can treat $\textrm{mean}(f_\textbf{j} (\textbf{x}) f_\textbf{k}(\textbf{y}))$ separately - for $H_0$ hypothesis of independence, all $a_{\textbf{j},\textbf{k}}$ should be from approximately normal distribution centered in 0.

To test it, let us first normalize them as in CLT, so that \textbf{for $H_0$ independence hypothesis, all $\hat{a}_{\textbf{j},\textbf{k}}$ should be from  $N(0,1)$}:
\be M=|B^+_x||B^+_y|\qquad \textrm{of}\qquad\hat{a}_{\textbf{j},\textbf{k}}= \frac{\textrm{mean}(f_\textbf{j} (\textbf{x}) f_\textbf{k}(\textbf{y}))}
{\sqrt{\textrm{var}(f_\textbf{j} (\textbf{x}) f_\textbf{k}(\textbf{y}))/n}}\label{norma}\ee

Assuming $H_0$ hypothesis, for pairwise dependencies: between one coordinate of each samples, due to independence and normalization above '$\textrm{var}$' variance has to be 1, allowing to omit the above estimation from sample. However, for higher order dependencies it does not have to be true due to dependencies inside samples - requiring to estimate variance from sample.

For a large $M=|B^+_x||B^+_y|$ number of features we can use standard tests if they are from $N(0,1)$ distribution like  Anderson–Darling~\cite{AD}, Shapiro-Wilk~\cite{SW}, $\chi^2$ or permutation tests. Below are proposed additional, in experiments the last one has provided the best sensitivity, hence is finally used.
\paragraph*{Log-likelihood test} of $\textrm{mean}(\ln(\textrm{PDF}(\hat{a}_{\textbf{j},\textbf{k}})))$, which assuming $H_0$ and large $M$ is approximately from $N(0,1)$, for which we get $E[\ln(\textrm{PDF})]=-(1+\ln(2))/2\approx -1.4189$ and variance 1/2, leading to test:
\be \textrm{Pr}(H_0)= 1- \textrm{erf}\left(\frac{|\textrm{mean}(\ln(\textrm{PDF}(\hat{a}_{\textbf{j},\textbf{k}})))+1.4189|}{2\sqrt{M}}\right)\ee

\begin{figure}[t!]
    \centering
        \includegraphics[width=90mm]{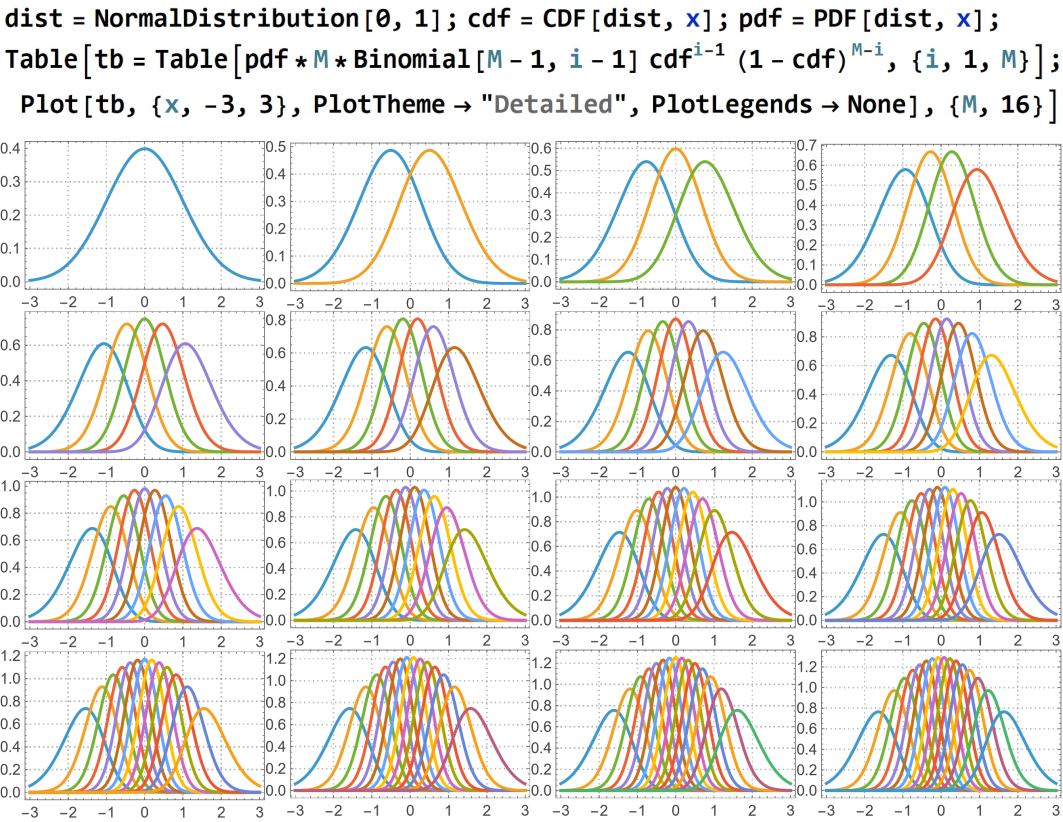}
        \caption{$\textrm{PDF}^M_i$ probability densities of sorted $M=1,\ldots,16$ variables from $N(0,1)$ normalized Gaussian distribution (analogously for different distributions), which can be used for tests if estimated normalized coefficients are from $N(0,1)$ for $\mathcal{H}_0$ hypothesis of independence, especially for the extreme values indicating the main dependence direction. As in Fig. \ref{svd}, it also allows to estimate significance of individual contributions, e.g. for joint distribution model of chosen significance level. 
        }
        \label{ordered}
\end{figure}

\paragraph*{Testing extremes $\{\hat{a}_{\textbf{j},\textbf{k}}\}_{\textbf{j}\in B_x^+,\textbf{k}\in B_y^+}$} for dependent samples, there might be some dominant dependence direction in joint distribution, which if covered in $M=|B^+_x||B^+_y|$ tested features, could dominate in estimated $\{\hat{a}_{\textbf{j},\textbf{k}}\}$, suggesting to focus on one or a few extreme values in both directions.

For hypothesis testing, we need to find \textbf{distribution of sorted $M$ values} from independent $N(0,1)$ or a different single variable distribution having given $\textrm{PDF}=\textrm{CDF}'$, e.g. for Marchenko-Pastur for SVD optimization. 
To find $\textrm{PDF}_{M,i}(x)$ density of $i$-th in order, there are $i-1$ smaller or equal, and $M-1-i$ greater then equal, combinatorially leading to $M {M-1 \choose i-1}$ possibilities, with density, shown in Fig. \ref{ordered}:
\be \textrm{PDF}^M_i(x)=M{M-1 \choose i-1}\, \textrm{PDF}(x)\, \textrm{CDF}(x)^{i-1}(1-\textrm{CDF}(x))^{M-i}
\ee
for $\textrm{PDF}\equiv \textrm{PDF}^1_1$, $\textrm{CDF}\equiv \textrm{CDF}^1_1$ of $N(0,1)$. By integrating it we can get its CDF for values from $N(0,1)$:
\be \textrm{CDF}^M_i(x)=M{M-1 \choose i-1}\, \textrm{Beta}(\textrm{CDF},i,1-i+M) \label{cdfm}
\ee
using incomplete Euler Beta function. We can use $\textrm{CDF}^M_i$ to test sorted values for hypothesis of being from a given distribution, also find those disagreeing for a given significance level.

\subsection{Chosen statistical significance joint distribution model}
\begin{figure}[t!]
    \centering
        \includegraphics[width=90mm]{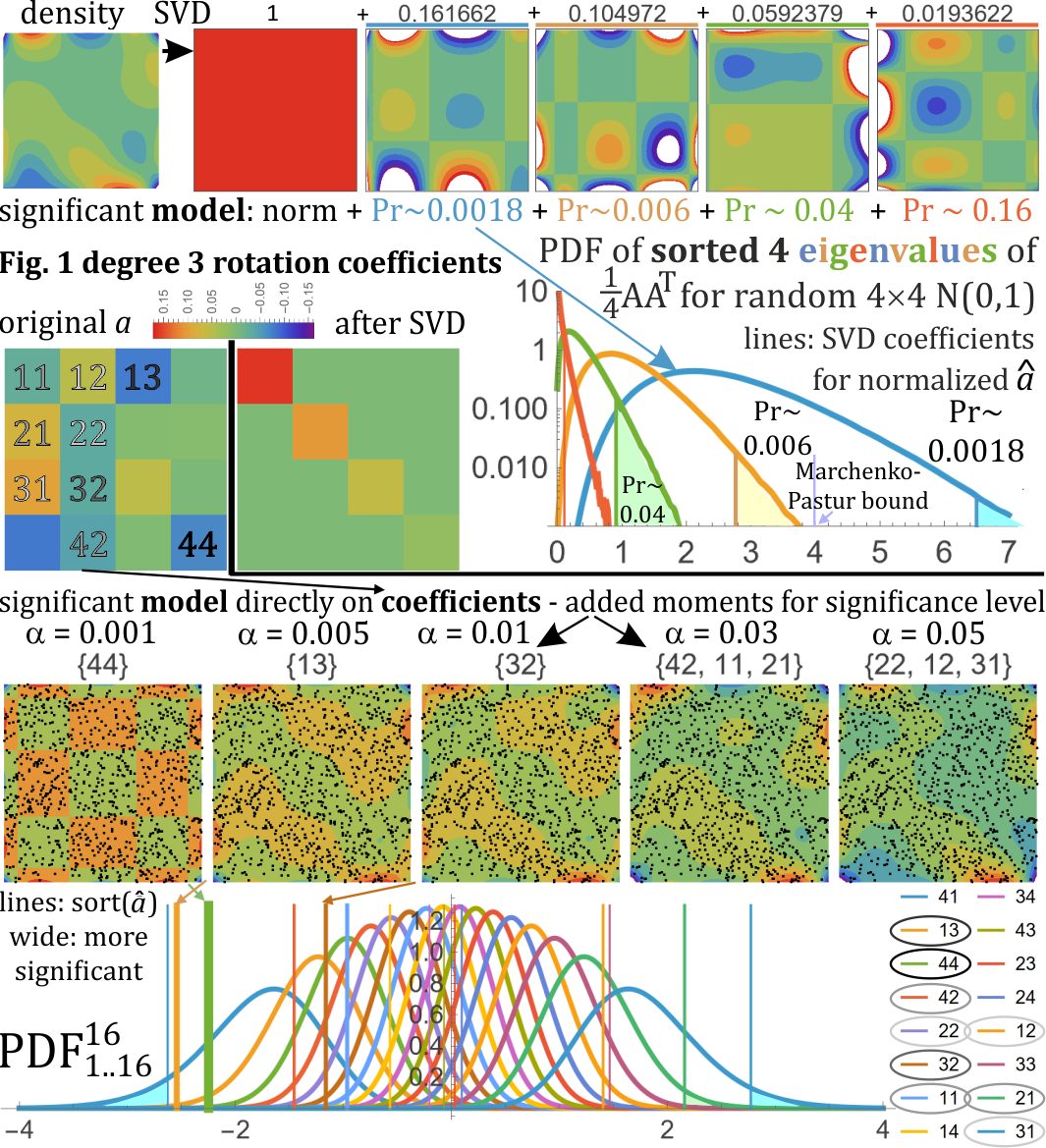}
        \caption{\textbf{Top}: we can perform SVD (singular value decomposition) of $|B_x^+|\times |B_y^+|$ matrix of coefficients, allowing to optimize their bases (\ref{basopt}), leading to decomposition of modeled joint density into contributions of controlled significance, allowing to use the highest or a few  below chosen significance level $\alpha$, here leading to similar Pr(independent) as HCRboth in Fig. \ref{intr}. To  distinguish signal from noise, we could use random matrix theory for singular values of $|B_x^+|\times |B_y^+|$ matrix with $N(0,1)$ random coefficients. Marchenko-Pastur theorem~\cite{MP} handles such situations, however, for the limit of infinite size, e.g. bounding singular values by 4.      
        In contrast, the plots shows densities from $10^6$ generated $4\times 4$ random matrices of $N(0,1)$ coefficients, clearly exceeding this bound - we can use such empirical estimation for practical approximation. It allows for joint distribution models using statistically significant contributions, e.g. for 0.01 significance we should take normalization plus the first two. 
        \textbf{Bottom}: analogously directly using coefficients  for various significance level using (\ref{sig}) formula. There are also shown $\textrm{PDF}^M_i$ densities for  $M=16$ sorted $N(0,1)$ values, and sorted $\hat{a}$ coefficients being $N(0,1)$ for independence hypothesis, showing that non-extreme values can also be significant.
        }
        \label{svd}
\end{figure}

Having $M$ estimated normalized coefficients $\hat{a}_{\textbf{j},\textbf{k}}$, assuming $H_0$ independence they should be from $N(0,1)$. We can evaluate  statistical significance of disagreeing with $H_0$ by individually comparing sorted normalized coefficients: $\textrm{sort}(\hat{a})$ with the above $\textrm{CDF}^M_i$ formula (\ref{cdfm}). It e.g. allows to build joint distribution model for a chosen significance level $\alpha$ - accepting only coefficients satisfying: 

\be\min\left(\textrm{CDF}^M_i(\textrm{sort}(\hat{a})_i),\ 1- \textrm{CDF}^M_i(\textrm{sort}(\hat{a})_i)\right)<\alpha \label{sig}\ee
for $N(0,1)$ distribution in joint density formula (\ref{jd}), or Marchenko-Pastur-like for SVD optimized as in Fig. \ref{svd}.

\subsection{Choice of basis $B$}
For $d_x=d_x=d=1$ dimensional samples, a natural choice of basis is just $B^+=\{1,2,\ldots,m\}$ for some $m$ number of moments, e.g. for $m=1$ we have only $a_{1,1}$ corresponding to just testing correlation. For $m=2$ we would additionally include homoscedasticity as $a_{2,2}$, but also $a_{1,2}$ and $a_{2,1}$ dependencies between expected value and variance, and so on, generally using $m^2$ features describing dependencies between data samples. 

Examples in this article use $m=4$ up to kurtosis with $m^2=16$ features, which through cross-validation usually turn out optimal e.g. for financial data as in Fig. \ref{global}, can be optimized for some specific data this way, or e.g. trying to recognize small rotations like in Fig. \ref{intr}.

For larger $d\geq 1$ dimensions it usually should be sufficient to focus on $(md)^2$ \textbf{pairwise dependencies} for bases with single nonzero index: 
\be \textrm{pairwise:}\quad B=\left\{\textbf{j}\in \{0,\ldots,m\}^d:\sum_i \textrm{sign}(j_i)=1\right\} \ee 
for fixed degree $m$, e.g. $m=1$ would correspond to testing only $d^2$ correlations between all pairs of coordinates. 

While we have discussed $d_x=d_y=d$, the two samples can be of different dimension, where we still can calculate mixed moments between their coordinates, estimate mutual information with (\ref{mif}) formula using different bases. Also, while for simplicity we have discussed fixed $m$, it might be worth to vary it between coordinates, maybe also change the basis e.g. to Fourier for periodic coordinates like day of year.

Dependence between samples should be usually detectable through such pairwise dependencies. However, in theory there can be only more subtle \textbf{higher order dependencies} - HCR independence test would need to consciously include to be able to detect, evaluate, describe. For example by adding to $B_x$ and/or $B_y$ basis $\sum_i \textrm{sign}(j_i)=2$ indexes, e.g. $a_{11,1}$ would include triplewise dependence like in Fig. \ref{basis}: that with change of two coordinates, there is change of expected value of the third one. The cost is larger number of features to estimate and analyze: $O(d^2)$ for pairwise, $O(d^3)$ for triplewise, and so on.

\subsection{PCA, CCA, SVD optimizations}
We can also perform some preprocessing of data to try to emphasize dependencies, hopefully making it more likely for low order e.g. pairwise test to find them, like PCA (principal component analysis) - rotation of coordinates to eigenvectors of correlation matrix, separately for both samples. 

PCA optimization often amplifies noise, e.g. Canonical Correlation analysis (CCA) allows to repair it - optimizing basis to maximize correlations, as discussed for HCR in \cite{cond3}. For independence test we could use it to maximize correlations between $f_\textbf{j}(\textbf{x})$ and $f_\textbf{k}(\textbf{y})$ values in bases for both samples.\\

For final $a_{\textbf{j},\textbf{k}}$ as $|B^+_x| \times |B_y^+|$ matrix we can also use SVD (singular value decomposition) into $U\Sigma V^T$ for orthogonal $U^T U=V^T V=I$ and diagonal $\Sigma$ of singular values, like in Fig. \ref{svd} we can optimize both used bases $f_{B_x}(\textbf{x})=(f_{\textbf{j}}(\textbf{x}))_{\textbf{j}\in B_x}$, $f_{B_y}(\textbf{x})=(f_{\textbf{k}}(\textbf{y}))_{\textbf{k}\in B_y}$ multiplying them by $U$ and $V$, to represent joint density:
\be \rho(\textbf{x},\textbf{y})=(f_{B_x}(\textbf{x}))   (a_{\textbf{j},\textbf{k}}) (f_{B_y}(\textbf{y}))^T=
\left((f_{B_x}(\textbf{x}))U\right) \Sigma \left((f_{B_y}(\textbf{y})) V\right)^T \label{basopt}\ee

We can freely perform such optimizations working on a family of data samples of similar behavior, using PCA, CCA or SVD of e.g. averaged samples, matrices. However, performing such optimization separately on a single tested sample would also amplify its noise - like in Fig. \ref{svd}, it needs modification of independence test to include it, e.g. based on random matrix theory, or synthetic empirical simulations.

\subsection{HCR independence test used in benchmarks}
The benchmarks for HCR have used basic pairwise dependencies for fixed degree $m=4$, testing minimal and maximal considered $\hat{a}$, presented as Algorithm \ref{test} and in Fig. \ref{intr}.

\begin{algorithm}[h!]
\normalsize{
\caption{\textbf{HCR basic pairwise independence test}()}
\label{test}
\begin{algorithmic}
\STATE $n, d_x, d_y, m\in \mathbb{N}$ \qquad \COMMENT{sample dimensions, chosen degree}
\STATE $U=\textrm{EDF}(X)$\qquad \qquad\COMMENT{coordinate-wise normalizations}
\STATE $V=\textrm{EDF}(Y)$\qquad \qquad\COMMENT{separately sorting coordinates}
\STATE $\hat{a}_{j,q,k,r}=\sqrt{n}\ \textrm{mean}(f_j(u_q)f_k(v_r))$ \quad \COMMENT{normalized features}
\STATE $\qquad $ for $q=1,\ldots,d_x, r=1,\ldots,d_y,$ and $j, k=1,\ldots,m$
\STATE $M=d_x d_y m^2$ \qquad\qquad\qquad \COMMENT{number of features}
\RETURN$\textrm{Pr}(\textrm{ind})=\textrm{CDF}^M_1(\min(\hat{a}))\, (1-\textrm{CDF}^M_M (\max(\hat{a})))$
\end{algorithmic}
}
\end{algorithm}
\noindent Accepting $H_0$ hypothesis if returned $\textrm{Pr(independent)}$ is above some chosen significance level $\alpha$, or reject otherwise.

It uses $\textrm{var}=1$ in (\ref{norma}) as mentioned allowed for pairwise dependencies, for triplewise and higher $\hat{a}$ should be additionally divided by estimated variance for coefficients $\textbf{j}, \textbf{k}$ having two or more nonzero indexes. For pairwise there is single nonzero index - we have denoted its position as $q$ and $r$. 

Its modifications can vary $m$ and $f$ bases between coordinates, add triplewise and higher order dependencies, include basis optimizations. 

It is also worth to consider different final tests if all $M$ normalized features are from $N(0,1)$ distribution - testing min and max is very sensitive to dependencies, but could lead to false positives, and ignores intermediate values. As we can see in Fig. \ref{svd}, also non-extreme values can be valuable to distinguish signal from noise, suggesting to \textbf{extend independence test to even more sensitive}: by using $2k\leq M$ extreme values, $i$-th sorted $\hat{a}$ as $\textrm{sort}(\hat{a})_i$ for $\textrm{Pr}(\textrm{independent})\approx $
\be 2^{2k} \prod_{i=1}^k \textrm{CDF}_i^M \left(\textrm{sort}(\hat{a})_i\right)\, \left(1-\textrm{CDF}_{M+1-i}^M \left(\textrm{sort}(\hat{a})_{M+1-i}\right)\right) \label{sorttest}
\ee
Where $2^{2k}$ is rescaling to get approximately 1 for independent. Analogously, as in Fig. \ref{svd}, we can perform such independence test after SVD basis optimization, comparing singular values of $\hat{a}_{\textbf{j},\textbf{k}}$ instead of the original coefficients.

To include all the values we could also use e.g. Anderson-Darling, Shapiro-Wilk, $\chi^2$, permutation, or log-likelihood test. 
\subsection{Toward kernel methods: global HCR vs local HSIC basis}
As in \cite{HCRNN}, we can represent the features as $n\times |B|
$ matrices:
\be \bar{X}=\frac{1}{\sqrt{n}} \left(f_\textbf{j}(\textbf{x})\right)_{\textbf{x}\in \hat{X},\, \textbf{j}\in B_x^+}\qquad
\bar{Y}=\frac{1}{\sqrt{n}} \left(f_\textbf{k}(\textbf{y})\right)_{\textbf{y}\in \hat{Y},\, \textbf{k}\in B_y^+}\ee
This way MSE estimation becomes $a_{\textbf{j},\textbf{k}}=\bar{X}^T\bar{Y}$ matrix, allowing to approximate mutual information (as trace is cyclic):
\be I(X,Y)\approx \textrm{Tr}(\bar{X}^T\bar{Y} (\bar{X}^T\bar{Y})^T)=\textrm{Tr}(K^X K^Y)\ee
$$\textrm{for }n\times n:\qquad K^X= \bar{X} \bar{X}^T\qquad K^Y= \bar{Y} \bar{Y}^T$$
kernel matrices we will also use for HSIC in the next section, with included correction like (\ref{mifc}) variance subtraction. 

However, while in HCR we focus on global basis of e.g. polynomials or Fourier, for HSIC there is usually used local basis like Gaussians, leading to essentially different types of represented density models, like in Fig. \ref{global}, with different behavior:
\begin{itemize}
    \item global basis is usually better at generalization: provides higher log-likelihood in cross-validation, as describing general features, instead of just assuming that new points will be close to the old points for local basis,
    \item global basis often allows to describe density with a reasonable number of features (moments), allowing more practical description e.g. for the discussed reduced complexity,
    \item local basis requires to choose kernel width, while global to choose basis with often universal e.g. $m=4$ degree,
     \item global orthonormal basis for normalized variables can work on deformation of $\rho=1$, allowing for practical approximations of entropy, mutual information using $\ln(1+t)\approx t$.
    \item above HCR kernel matrices include all $|B_x^+| |B_y^+|$ features, while for HSIC Gaussian kernel only single $\sigma$ is used: 
\end{itemize}
$$K^\textrm{HCR}_{\textbf{x},\textbf{x}'}=\frac{1}{n}\sum_{\mathbf{j}\in B_x^+,\mathbf{k}\in B_x^+} f_\textbf{j}(\textbf{x})f_\textbf{k}(\textbf{x}')\quad \qquad K^\textrm{HSIC}_{\textbf{x},\textbf{x}'}=e^{-\frac{ \|\textbf{x} - \textbf{x}'\|_2}{2\sigma^2}}$$
maybe worth to extend e.g. to multiple $\sigma_i$: $\sum_i  \exp\left(-\frac{ \|\textbf{x} - \textbf{x}'\|_2}{2\sigma_i^2}\right)$.

\section{Hilbert--Schmidt Independence Criterion (HSIC)}

\subsection{Statistical dependence measures}

Let $(X,Y)$ be random variables with joint distribution $\mathcal R$ on $\mathcal X\times\mathcal Y$.
A (population) dependence measure is a functional
\[
\Psi:\ \mathcal{P}(\mathcal X\times\mathcal Y)\to[0,\infty).
\]
Following classical desiderata (e.g., Rényi), one may require (see \cite{kernelmethods}):
\begin{itemize}
    \item $\Psi$ is well-defined
    \item $0 \leq \Psi(\mathcal{R}) \leq 1$
    \item $\Psi(\mathcal{R}) = 0$ if and only if $X, Y$ are independent
    \item $\Psi(\mathcal{R}) = 1$ if and only if $Y = g(X)$ for some deterministic bijective $g$
\end{itemize}
These are optional properties and do not have to hold for every measure; 
in particular, HSIC is nonnegative and equals zero iff $X$ and $Y$ are independent 
(with characteristic kernels), but it is not generally bounded by $1$ and does not attain $1$ for deterministic dependence. 

A standard example of such statistical dependence measure in mutual information measuring the number of bits one variable says about the second, however, changing the upper bound from 1 to entropy for samples differing by bijection. In contrast to HCR, HSIC does not have such interpretation, however, is used as its practical replacement e.g. in information bottleneck training~(\cite{IB1,IB2}).

In practice we work with a sample 
$Z=\{(\textbf{x}^i,\textbf{y}^i)\}_{i=1}^n$ and a sample statistic
$\widehat{\Psi}:(\mathcal X\times\mathcal Y)^n\to\mathbb R$ used as a test statistic.

\subsection{Definitions}

\subsubsection{Reproducing Kernel Hilbert Spaces}

Let $\mathcal{X}$ be set the data comes from and $\mathcal{H}$ a Hilbert space. Let $\varphi: \mathcal{X} \rightarrow \mathbb{R}$ be the feature map, mapping datapoints to points in $\mathcal{H}$. 

We call $\kappa: \mathcal{X} \times \mathcal{X} \rightarrow \mathbb{R}$ a kernel if~\cite{kernelintro}
\[ \kappa(\textbf{x}, \textbf{y}) = \langle \varphi(\textbf{x}), \varphi(\textbf{y}) \rangle_{\mathcal{H}} \] 

We call $\kappa$ a reproducing kernel and the Hilbert space $\mathcal{H}$ of functions $\mathcal{X} \rightarrow \mathbb{R}$ a reproducing kernel Hilbert Space if the following conditions are satisfied for every $x \in \mathcal{X}$ and every $f \in \mathcal{H}$: 

\begin{enumerate}
    \item $\kappa(\textbf{x}, .) \in \mathcal{H}$
    \item $f(\textbf{x}) = \langle f, \kappa(\textbf{x}, .) \rangle_{\mathcal{H}}$
\end{enumerate}

The most popular kernel choice is the Gaussian function, so 

\[ \kappa(\textbf{x}, \textbf{y}) = e^{-\frac{||\textbf{x}-\textbf{y}||_2}{2 \sigma^2 }}\]

for some fixed variance $\sigma$. 

In this case $\mathcal{H}$ is the space of Gaussian mixtures and 

\[ \phi(\textbf{x}) = t \rightarrow e^{-\frac{||\textbf{x}-\textbf{t}||_2}{2 \sigma^2 }} \]

\subsubsection{Kernel Mean Embedding}

Let $\kappa$ be a reproducing kernel and $\mathcal{H}$ a RKHS. Let $\mathcal{P}: \mathcal{X} \rightarrow \mathbb{R}$ be a probability distribution on $\mathcal{X}$. Then, by the Riesz Representation Theorem, there exists an unique $\mu_P \in \mathcal{H}$ such that $\forall f \in \mathcal{H}$

\[ \langle f, \mu_P \rangle_{\mathcal{H}} = \mathbb{E}_{\textbf{x} \sim \mathcal{P}} (f(\textbf{x}))\]

It can be written as~\cite{kernelintro}

\[ \mu_P = \mathbb{E}_{x \sim \mathcal{P}} (\kappa (x, .)) \]

\subsubsection{Cross-Covariance Operator}

Let $X: \mathcal{X} \rightarrow \mathbb{R}, Y: \mathcal{Y} \rightarrow \mathbb{R}$ be two random variables. Let $\mathcal{F}, \mathcal{G}$ be their respective RKHS's with associated kernel functions $k, l$. Let $\mathcal{R}$ be the joint distribution of $X, Y$ and $\mathcal{P}$, $\mathcal{Q}$ associated marginal distributions. 

The cross-covariance operator $C_{\mathcal{R}}: \mathcal{F} \rightarrow \mathcal{G}$ is defined by satisfying the following condition for every $f \in \mathcal{F}, g \in \mathcal{G}$:

\[ \langle f, C_{\mathcal{R}} (g) \rangle_{\mathcal{F}} = \mathbb{E}_{\textbf{x}, \textbf{y}: \mathcal{R}} (f(\textbf{x})-\mathbb{E}_{\textbf{x}': \mathcal{P}} (f(\textbf{x}')))(g(y)-\mathbb{E}_{\textbf{y}': \mathcal{Q}} (f(\textbf{y}'))) \]

Which can be written as~\cite{kernelintro}

\[ C_{\mathcal{R}} = \mathbb{E}_{\textbf{x}, \textbf{y} \sim \mathcal{R}}(k(\textbf{x}, .)-\mu_{\mathcal{P}}) \otimes (l(\textbf{y}, .) - \mu_{\mathcal{Q}})) \]

where $\otimes$ denotes a tensor product of these two Hilbert spaces.

\subsubsection{Hilbert-Schmidt Norm}

Let $\mathcal{F}, \mathcal{G}$ be Hilbert spaces and $f \in \mathcal{F}, g \in \mathcal{G}$. Then the Hilbert-Schmidt norm of the tensor product $f \otimes g$ is defined as

\[ || f \otimes g ||_{HS} = ||f||_{\mathcal{F}} ||g||_{\mathcal{G}} \]

\subsection{HSIC Formula}

Value of the HSIC measure of the joint probability distribution $\mathcal{R}$ is the Hilbert-Schmidt norm of its covariance operator, so $||C_{\mathcal{R}}||_{HS}$. This can be further transformed into~\cite{kernelintro}:

\be \| C_{\mathcal{R}}\|_{HS} = \mathbb{E}_{\textbf{x}, \textbf{y}\sim \mathcal{R}, \textbf{x}', \textbf{y}' \sim \mathcal{R}(k(\textbf{x}, \textbf{x}')l(\textbf{y}, \textbf{y}')) + }\ee 
$$\mathbb{E}_{\textbf{x} \sim \mathcal{P}, \textbf{x}' \sim \mathcal{P} (k(\textbf{x}, \textbf{x}')) \mathbb{E}_{\textbf{y} \sim \mathcal{Q}, y' \sim \mathcal{Q} (l(\textbf{y}, \textbf{y}')) - 2 \mathbb{E}_{\textbf{x}, \textbf{y} \sim \mathcal{R}} (\mathbb{E}_{\textbf{x}' \sim \mathcal{P}} (k(\textbf{x}, \textbf{x}')) \mathbb{E}_{\textbf{y}' \sim \mathcal{Q}} (l(\textbf{y}, \textbf{y}')))}}$$


\subsection{HSIC Estimators}

In practice stastistical dependence tests are performed on a set of $n$ datapoints instead of the distribution formula. 
Let 
\[ Z = \{ (\textbf{x}^1, \textbf{y}^1), (\textbf{x}^2, \textbf{y}^2), \ldots, (\textbf{x}^n, \textbf{y}^n) \} \]
be the dataset. 

Let $K, L \in \mathbb{R}^{n \times n}$ be kernel matrices such that 
\be K_{ij} = k(\textbf{x}^i, \textbf{x}^j) \qquad L_{ij} = l(\textbf{x}^i, \textbf{x}^j)\ee
The Gaussian function is typically chosen as $k, l$. 

Let 
\[ H = I_n - \frac{1}{n} J_n\]
be the centering matrix where $J_n$ is an $n \times n$ matrix with all ones. 

Then the V-statistic for HSIC is~\cite{kernelintro, HSIC, kernelmethods}:

\[ \textrm{HSIC}(Z) = \frac{1}{n^2} \textrm{Tr}(KHLH) \]

\subsection{HSIC as a stastical test}

In order to turn HSIC into the statistical dependence test, we state the null hypothesis $H_0$ as $\mathcal{R} = \mathcal{P} \mathcal{Q}$ and the alternative hypothesis $H_1$ as $\mathcal{R} \neq \mathcal{P} \mathcal{Q}$. Given the test statistic $\textrm{HSIC}(Z)$ and the desired test level $\alpha$, null hypothesis is rejected if $\textrm{HSIC}(Z) > \theta$ for some threshold $\theta$ such that the probability of type I error is $\alpha$. 

The most popular method for deriving $\theta$ is approximating the distribution of $\textrm{HSIC}(Z)$ under $H_0$ with Gamma function~\cite{HSIC}

\[ \textrm{HSIC}(Z) \approx \frac{x^{a-1} e^{-\frac{x}{b}}}{b^a \Gamma(a)} \]
where
\[ a = \frac{\mathbb{E}(\textrm{HSIC}(Z))^2}{\textrm{var}(\textrm{HSIC}(Z))}\]
\[ b = \frac{n \cdot \textrm{var}(\textrm{HSIC}(Z))}{\mathbb{E}(\textrm{HSIC}(Z))} \]

Then $\theta$ is equal to its $1-\alpha$ quantile. 

\subsection{HSIC vs.\ HCR}

Both HSIC (Hilbert--Schmidt \emph{Independence} Criterion) and HCR (Hierarchical Correlation Reconstruction) produce sample statistics that increase with dependence, but they arise from different geometries and offer complementary trade-offs.

\paragraph*{Geometry and what is being measured}
HSIC operates in an RKHS induced by user-chosen kernels $k$ and $\ell$ on $\mathcal X$ and $\mathcal Y$. With a characteristic choice (e.g., RBF), HSIC is zero if and only if $X$ and $Y$ are independent. Equivalently, HSIC is the squared MMD between the joint and the product of marginals with product kernel $s((x,y),(x',y'))=k(x,x')\,\ell(y,y')$.  
HCR works in a global, orthonormal polynomial basis on the copula domain after marginal normalization (CDF/EDF). Mixed moments $a_{\mathbf j,\mathbf k}$ capture interpretable interactions (mean–mean, mean–variance, etc.). A practical MI proxy is the sum of squares of nontrivial mixed moments.

\paragraph*{“Local” vs.\ “global” sensitivity}
RBF kernels do not impose a KDE model per se; rather, the bandwidth $\sigma$ controls the effective locality of the RKHS features. For characteristic kernels, HSIC can capture global dependencies; poor bandwidth selection may reduce power to either very local or overly smooth effects.  
In HCR, the polynomial degree $m$ controls global complexity: higher $m$ adds higher-order mixed moments (e.g., skewness, kurtosis interactions). This often yields good out-of-sample generalization for global structure and offers straightforward interpretability.

\paragraph*{Computational profile}
A standard HSIC implementation forms $n\times n$ Gram matrices, incurring $O(n^2)$ time and memory. This is often the practical bottleneck for large $n$.  
HCR computes each feature in one pass, $O(n)$, so testing a set $B$ mixed-moment features costs $O(n\,|B|)$ time and $O(|B|)$ memory, which scales well to large $n$ (with linear streaming updates).

\paragraph*{Calibration and hyperparameters}
HSIC thresholds can be obtained via permutation or gamma moment-matching for the biased V-statistic; the main hyperparameter is kernel choice and bandwidth (often tuned by CV or median heuristic).  
HCR requires choosing the basis (e.g., Legendre) and degree $m$, and possibly which coordinates/pairs to include. Calibration can use normalization of features to $N(0,1)$ under $H_0$ and multiple-testing control (e.g., via the distribution of $\max_i |\hat z_i|$ or permutations).

\paragraph*{Interpretability and extensibility}
HSIC yields a single scalar without a direct decomposition into human-readable effects.  
HCR is explicitly decomposable: each coefficient corresponds to a mixed moment, enabling diagnosis of \emph{which} interactions drive dependence. Extending from pairwise to triplewise or higher-order interactions is straightforward by enlarging $B$ (with cubic and higher growth in $d$ if needed).

\paragraph*{When to prefer which}
For small/medium $n$ and when a powerful, kernel-agnostic test is sufficient, HSIC is a strong default.  
For very large $n$, streaming settings, or when interpretability and targeted feature testing matter, HCR is attractive due to linear cost and mixed-moment readouts. In practice, they are complementary: HSIC for broadly powered detection, HCR for scalable detection plus explanation of the dependence structure.


\section{Summary and further work}
There was proposed application of HCR mutual information evaluation for independence test, and comparison with state-of-art HSIC. In much lower complexity, finally practical to also work with very large data samples - crucial to find very subtle dependencies, it usually obtains even better sensitivity. Additionally, HCR provides evaluation of mutual information, and interpretable description of the found dependencies, allowing to model joint distribution between data samples, including significance evaluation of contributions. Finding dependencies, We can further exploit them to predict conditional densities - e.g. using HCR techniques from \cite{cond1,cond2,cond3}.

\begin{figure*}[h]
    \centering
        \includegraphics[width=190mm]{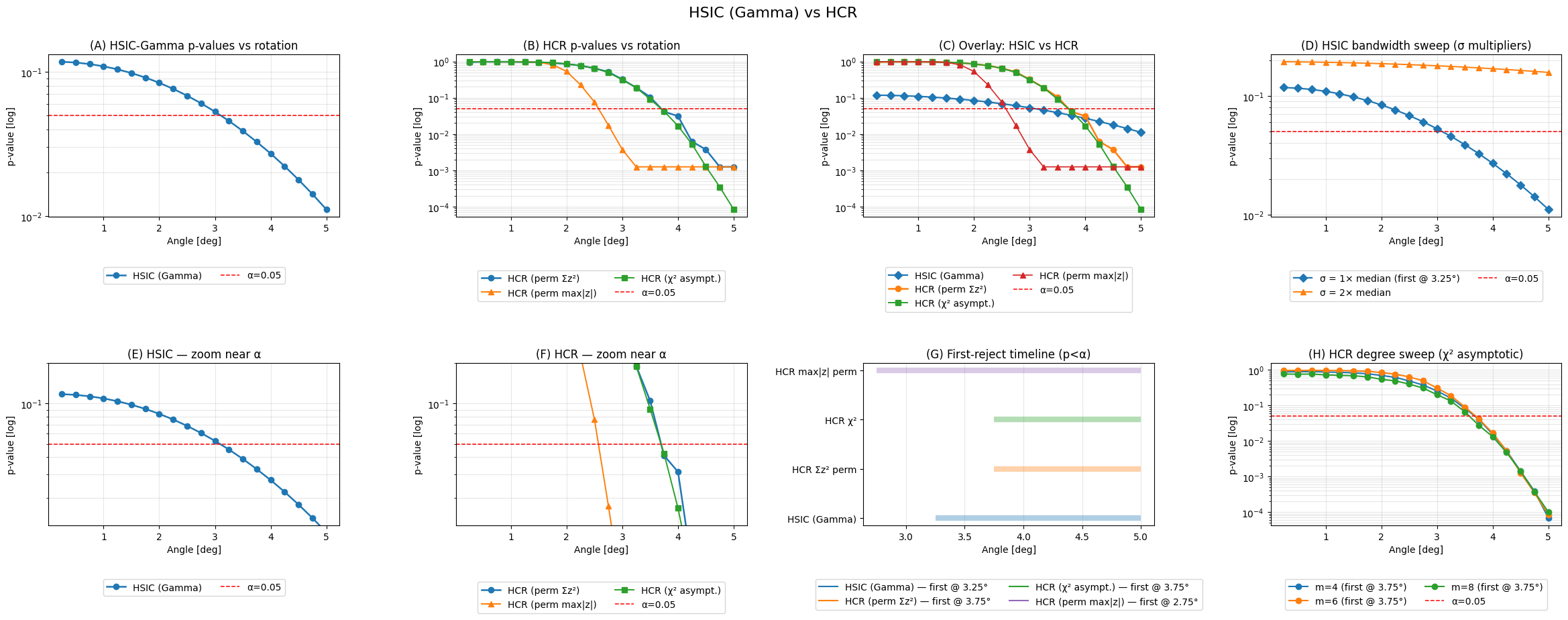}
        \caption{\textbf{HSIC (Gamma) vs HCR.}
        We start from $n=1000$ i.i.d.\ samples $(x_1,x_2,y_1,y_2)$ with $X=(x_1,x_2)$ independent of $Y=(y_1,y_2)$.
        A $50^\circ$ within–block rotation is applied to $(x_1,x_2)$ (which preserves independence), and then a fixed global rotation
        $R\in\mathbb{R}^{4\times4}$ is applied repeatedly to the whole 4D vector; each application adds $\Delta\theta=0.25^\circ$
        (the $x$-axis shows the cumulative angle).
        At each angle we test $H_0\!:\,X\perp Y$ at $\alpha=0.05$ and plot $p$-values on a log scale (red dashed line marks $\alpha$).
        \textbf{HSIC (Gamma)} uses Gaussian RBF kernels with bandwidths set by the median heuristic on $X$ and $Y$; the biased V-statistic is calibrated by a Gamma fit.
        \textbf{HCR} maps marginals to the copula domain via EDF and uses an orthonormal Legendre basis of degree $m$; we report
        (i) permutation $p$ for $T=\sum z^2$ (``$\Sigma z^2$ perm''),
        (ii) permutation $p$ for $\max|z|$ (``max$|z|$ perm''), and
        (iii) the $\chi^2$ approximation for $T$.
        Default settings: $m=6$, $B=800$ permutations.\\[2pt]
        \textbf{(A)} HSIC–Gamma $p$-values vs rotation; first rejection at $\approx 3.25^\circ$.
        \textbf{(B)} HCR $p$-values vs rotation for the three calibrations; max$|z|$ perm rejects earliest ($\approx 2.75^\circ$), $\Sigma z^2$ perm and $\chi^2$ cross later ($\approx 3.75^\circ$).
        \textbf{(C)} Overlay of HSIC–Gamma and HCR on the same samples; all methods agree and decay with angle.
        \textbf{(D)} HSIC bandwidth sweep: RBF $\sigma$ set to median vs $2\times$median; larger bandwidth is slightly more conservative, confirming the median heuristic as a reasonable default.
        \textbf{(E)} HSIC—zoom near $\alpha$ (detail of the crossing).
        \textbf{(F)} HCR—zoom near $\alpha$ for all three calibrations.
        \textbf{(G)} First-reject timeline ($p<\alpha$): horizontal markers summarize the angle of the first rejection for each method (earlier $\Rightarrow$ higher sensitivity).
        \textbf{(H)} HCR degree sweep (asymptotic $\chi^2$): comparison of $m\in\{4,6,8\}$ shows a modest gain from higher $m$ with stable behavior.
        Overall, increasing global rotation induces dependence and all tests detect it; HCR (max$|z|$ perm) is the most sensitive here, followed by HSIC–Gamma, while $\Sigma z^2$ perm and $\chi^2$ calibrations are slightly more conservative but track the same trend.}
        \label{fig:hsic-hcr}
\end{figure*}

As further work, the proposed tests likely can be improved, also from theoretical perspectives, choice of details especially of the final normality test - improving sensitivity and reducing false positive probability, e.g. Fig. \ref{fig:hsic-hcr} uses some standard ones. The proposed test (\ref{sorttest}) of sorting and individually comparing multiple values with distributions of sorted (\ref{cdfm}) e.g. $N(0,1)$ seems promising general approach for normality test, also for other distributions - worth to be explored.

There is also mentioned promising research direction of basis optimization e.g. to make it more likely for pairwise test to find dependencies - like initial PCA or CCA of data samples, or directly SVD of $a_{\textbf{j},\textbf{k}}$ matrix to optimize basis as in Fig. \ref{svd}. It also amplifies noise - should be used on a family of similar samples to optimize basis for the given task, or for single sample using modified independence test e.g. based on random matrix theory or simulations randomly generating matrices.

Finally we should explore applications, like finding and exploiting subtle dependencies e.g. in financial data, or information bottleneck neural network training. Also compare with state-of-art mutual information estimation like \cite{MI}.\\

\textbf{Author contributions}: Jagoda Bracha and Adrian Przybysz are responsible for Section III about HSIC and its tests with library, Jarek Duda is responsible for the rest.\\

\textbf{Acknowledgments}: The presented research was carried out within the AIntern program  (\url{https://www.qaif.org/events/aintern}) organized by Fundacja Quantum AI, for project "Biology-inspired artificial neurons with a joint distribution model".

\bibliographystyle{IEEEtran}
\bibliography{cites}

\end{document}